\documentclass[10pt,reqno]{amsart}

\usepackage{graphicx}
\usepackage{amsmath,amssymb}
\usepackage{natbib}
\usepackage{booktabs}
\usepackage{multirow}
\usepackage[hidelinks]{hyperref}

\newcounter{algorithm}
\renewcommand{\thealgorithm}{\arabic{algorithm}}
\newcommand{\method}{ConceptADapt}

\makeatletter
\newcommand{\tablecaption}{\def\@captype{table}\caption}

\newcommand{\paperaffiliation}[1]{\gdef\@paperaffiliation{#1}}
\newcommand{\paperemail}[1]{\gdef\@paperemail{#1}}
\let\ams@setauthors\@setauthors
\renewcommand{\@setauthors}{%
  \ams@setauthors
  \vspace{-0.4em}
  \begin{center}
    \footnotesize\normalfont
    \@paperaffiliation\\[0.15em]
    \href{mailto:\@paperemail}{\texttt{\@paperemail}}
  \end{center}
  \vspace{0.5em}
}
\makeatother

\title[ConceptADapt for Few-Shot Industrial Anomaly Detection]{\method: Concept-guided Adaptive Feature Reconstruction with Dynamic Attention for Few-Shot Industrial Anomaly Detection}
\author{Yufei Li, Yicheng Ruan, Long Tian, Dongsheng Wang, Liang Bao}
\paperaffiliation{School of Computer Science and Technology, Xidian University, Xi'an, China}
\paperemail{tianlong@xidian.edu.cn}
\date{}

\begin{document}

\begin{abstract}
Few-shot industrial anomaly detection (FS-IAD) focuses on detecting and localizing visual defects in industrial inspection during the cold-start phase, where only a limited number of normal training samples are available per category. Recent advances in this field predominantly leverage visual features from foundation-model and have achieved promising performance. Despite the strong representational power of foundation-model features, the model generalization remains fragile due to the extreme scarcity of normal training data.
To address this pivotal issue, we propose ConceptADapt, a concept-guided adaptive feature reconstruction model with dynamic attention. Specifically, our model pre-learns a set of fixed normal concepts from the limited support features and leverages them to mine relationships with query features, thereby recalibrating their statistics for improved anomaly detection at test time. To mitigate the prevalent feature shortcut problem, which is particularly severe under low-data regimes, we further develop a dynamic attention mechanism integrated with sparse autoencoders to learn robust normal concepts during training. Moreover, to enable fast adaptation during inference, our model remains lightweight by incorporating LoRA into the attention module, which introduces only minimal updating parameters.
Extensive experiments on three widely adopted FS-IAD benchmarks, including MVTec-AD, VisA, and MPDD, demonstrate that our model consistently outperforms state-of-the-art (SOTA) approaches across both detection and localization tasks, achieving significant improvements under various shot settings.
\end{abstract}

\maketitle

\section{Introduction}

Industrial anomaly detection (IAD) aims to identify visual defects based on knowledge learned from normal training samples. It plays a vital role in the product manufacturing pipeline, as even subtle defects can pose safety risks or cause downstream task failures. In practice, however, achieving precise and reliable IAD is challenging due to the rarity, diversity, and unpredictability of anomalies, which are often unseen during training \cite{mvtec, roth2022patchcore}. Furthermore, the cold-start problem for new products exacerbates this difficulty, as normal training samples become scarce, giving rise to the few-shot IAD (FS-IAD) setting \cite{visa, regad}. Consequently, developing robust and automated detection systems for FS-IAD has attracted increasing research attention.

\begin{figure}[t]
\centering
\includegraphics[width=\columnwidth]{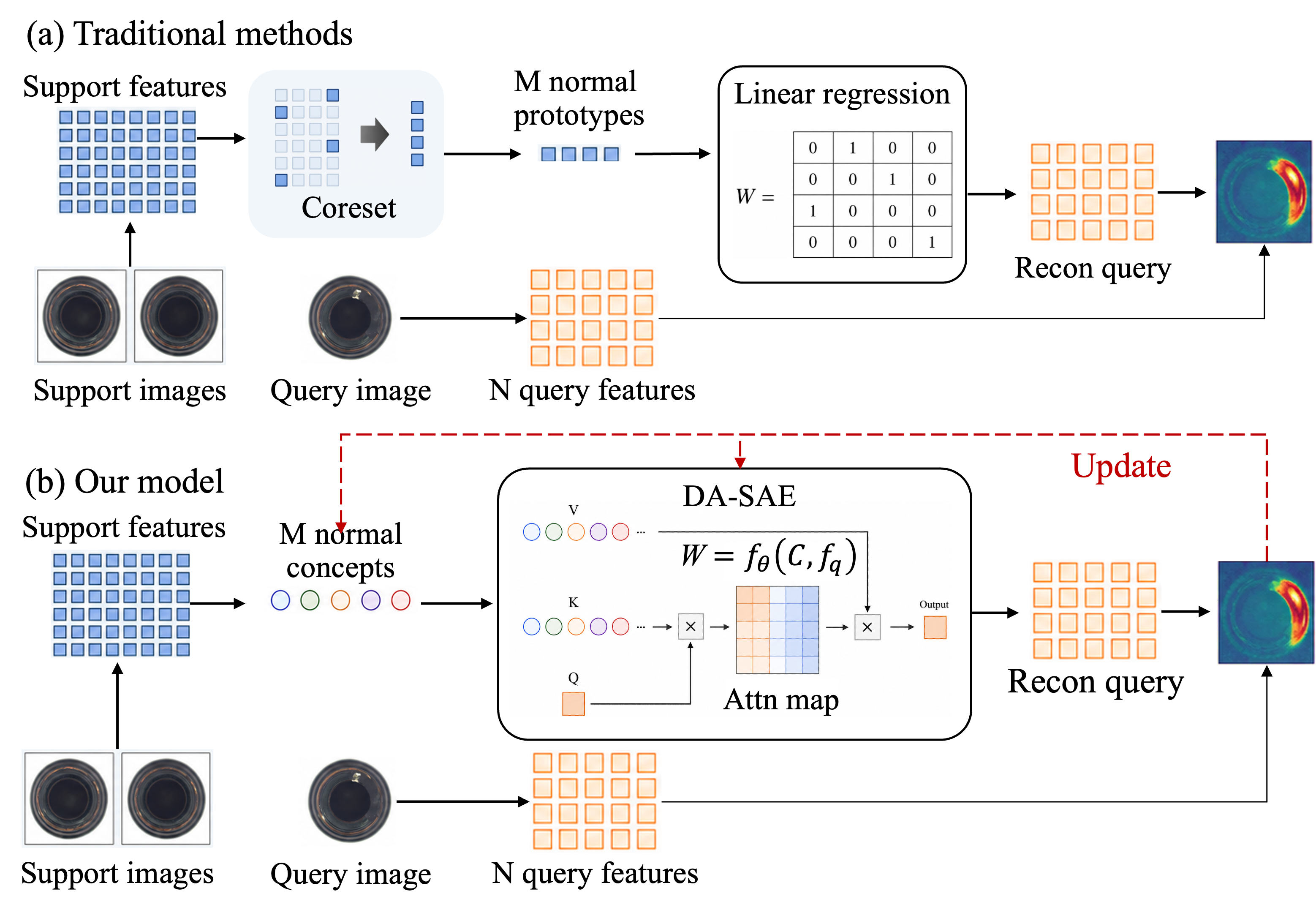}
\caption{Comparison between our method and conventional prototype-based approaches. (a) Conventional methods typically compress support features into a prototype set via Coreset \cite{agarwal2005geometric}. Anomaly detection is then performed by comparing query features with their reconstructions, which are obtained as a linear combination of the normal prototypes. However, this strategy is susceptible to false responses due to incomplete prototype coverage and the shortcut learning. (b) In contrast, our model learns a set of normal concepts using a dynamic attention-modulated sparse autoencoder (DA-SAE), effectively circumventing shortcut issues under low-data regimes. At inference, we further employ concept-guided adaptive feature reconstruction, which adapts the query feature statistics to the DA-SAE via LoRA.}
\label{fig:mechanism}
\vspace{-2mm}
\end{figure}

To address the few-shot challenge of IAD, meta learning-based methods \cite{regad, wu2021learning} and metric-based methods \cite{roth2022patchcore, jeong2023winclip, fastrecon,foct,damm2025anomalydino} have been explored, respectively. 
For example, to alleviate the adverse effects of limited normal training samples,
\cite{regad} employs registration, an inherently category-generalizable image alignment task, as the proxy task, to train a category-agnostic anomaly detection model. Meanwhile, \cite{damm2025anomalydino} follows the patch-level deep nearest neighbor paradigm and leverages foundation-model features, such as those from DINOV2 \cite{oquab2023dinov2}, to enhance detection under few-shot conditions. Go one step further, \cite{fastrecon} and \cite{foct} propose to transfer query feature statistics into the normal prototypes learned from few training samples via regularized ridge regression and optimal transport \cite{cuturi2013sinkhorn}, respectively. 
Despite their effectiveness, these methods suffer from two notable shortcomings. First, they rarely account for the feature shortcut problem \cite{you2022unified}, where the model tends to recover anomalies due to identity mapping caused by shortcut learning during training, an issue that we argue becomes more pronounced when normal samples are scarce. Second, existing metric-based methods that transfer query statistics often require adjusting the learned normal prototypes, which may inadvertently introduce additional noise.

In this work, we present \method{}, a concept-guided adaptive feature reconstruction model with dynamic attention, to tackle the two aforementioned obstacles in a unified framework. As illustrated in Fig. \ref{fig:mechanism}, to address the feature shortcut problem, we propose a dynamic attention-modulated sparse autoencoder (DA-SAE) built upon foundation-model features. The DA-SAE learns both the normal concepts and the decoder-only parameters, where the normal representations are absorbed in concepts by the learned attention patterns. We observe that the sparsity induced by the dynamic attention is beneficial for mitigating shortcut learning, serving as a systematic and mathematically principled alternative to the Dropout operation commonly used in conventional Transformer block. To circumvent the prototype-shift problem during query feature transfer, we freeze the learned concepts and employ a one-layer decoder, updating only the last feed-forward network (FFN) layer via LoRA \cite{hu2022lora}. This design not only prevents prototype drift but also enables fast adaptation during inference. To further improve efficiency and reduce the risk of introducing noise, we re-weight the feature reconstruction loss using the inverse of the anomaly score, thereby encouraging the model to focus on normal regions while down-weighting potential anomalous areas. Our contributions are:
\begin{itemize}
    \item We propose \method{}, a concept-guided adaptive feature reconstruction model with dynamic attention, to address the long-standing issues of feature shortcut and prototype shift in a unified framework.
    \item We design a lightweight DA-SAE decoder that enables fast query feature transfer while minimizing the risk of introducing anomalies at test time.
    \item We conduct comprehensive experiments on MVTec-AD, VisA, and MPDD. The results demonstrate that our model achieves competitive or SOTA performance on both detection and localization tasks across various shot settings.
\end{itemize}

\section{Related Work}

\subsection{Industrial Anomaly Detection}
Industrial anomaly detection (IAD) aims to identify product defects by learning a compact normal boundary within the normal feature manifold \cite{defard2021padim,liu2023simplenet,batzner2024efficientad}, typically leveraging abundant normal training samples. Existing methods primarily focus on mitigating the feature shortcut learning problem to obtain a reliable normal manifold for improved detection \cite{zavrtanik2021draem,deng2022reverse}, and can be broadly categorized into two groups. The first line of work \cite{roth2022patchcore, hvqtrans} mitigates shortcut learning from the feature perspective by constructing discrete normal prototypes. The second line \cite{you2022unified, deng2024dinomaly} employs Transformer-based cross-attention or self-attention mechanisms to circumvent shortcut learning.

The feature shortcut problem is equally critical and may become even more severe in low-data IAD scenarios. However, this issue has been rarely studied in existing work, and our model is specifically designed to address this direction.

\subsection{Few-shot Industrial Anomaly Detection}
In contrast to standard IAD, few-shot IAD (FS-IAD) is characterized by severely limited normal training data, which leads to less representative prototypes. Consequently, conventional IAD approaches become inadequate in this setting \cite{santos2023fspatchcore}. To improve prototype representativeness, two main research lines have been explored. The first line \cite{jeong2023winclip, damm2025anomalydino, subspacead} typically relies on visual foundation models \cite{oquab2023dinov2} or multi-modal tuning of vision-language models \cite{radford2021learning} to enhance the manifold representation, thereby indirectly improving the representational capacity of the prototypes. The second line \cite{fastrecon, foct} also builds upon pre-trained \cite{tan2019efficientnet, he2016deep} or foundation models, and further calibrates prototypes with various metrics by transferring the statistics of query features through mining the relationships between the original prototypes and online query features.

Calibrating prototypes based on the relationships between original prototypes and query features may induce prototype shift, which in turn introduces anomalies. To mitigate this issue, various regularizations and hyperparameters are typically introduced, inevitably increasing model complexity. Our work aims to address this remaining conflict.

\begin{figure}[t]
\centering
\includegraphics[width=1.\textwidth]{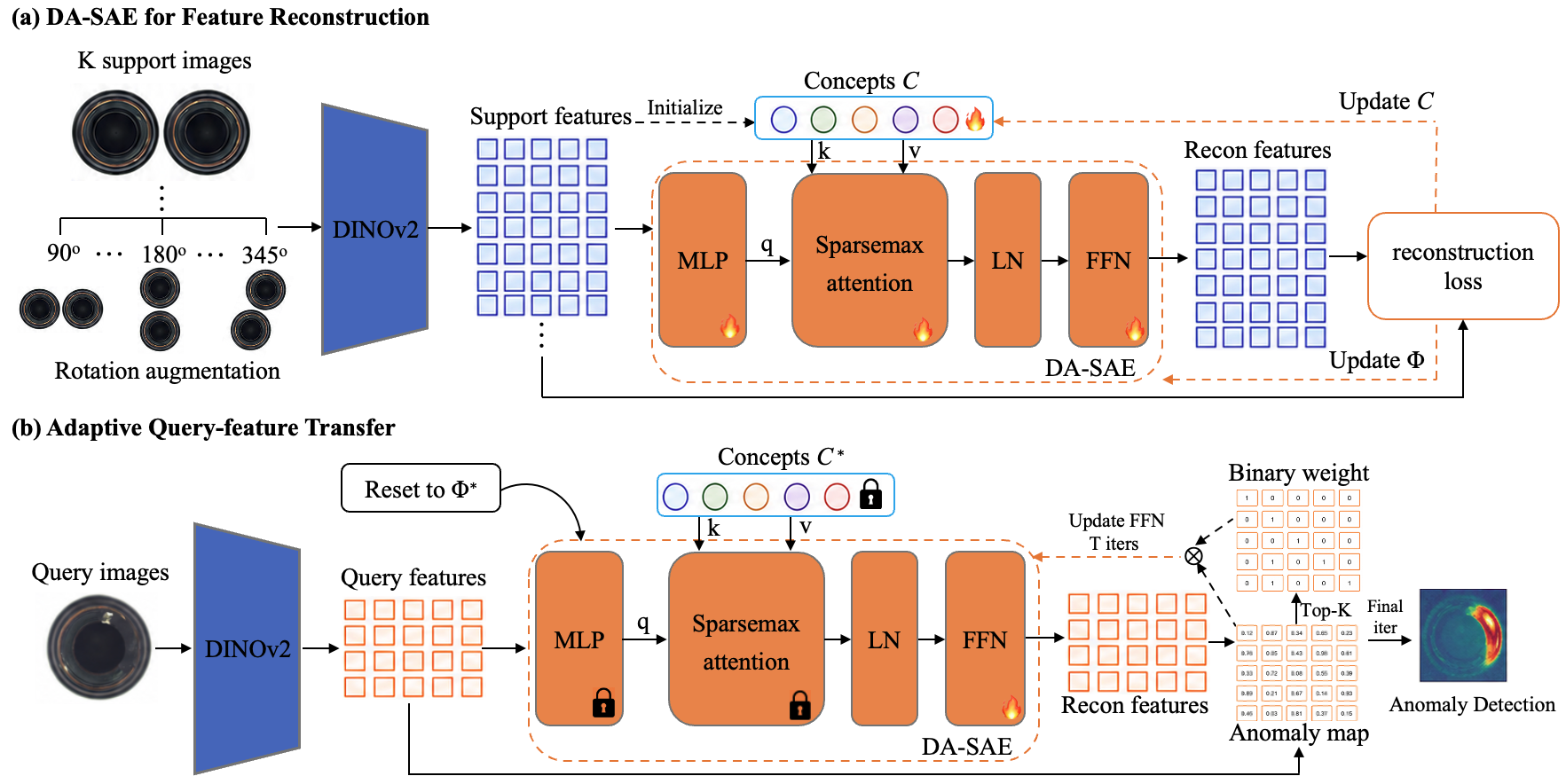}
\caption{Overview of our \method{}. During training (top), we first extract augmented support features using DINOv2, and then optimize the DA-SAE and learnable concepts via feature reconstruction. At inference (bottom), we freeze the learned concepts and update only the last FFN layer of the DA-SAE using LoRA, with feature reconstruction as the supervisory signal. To further suppress the potential anomalous patterns, we re-weight the reconstruction loss by the inverse of the anomaly score.}
\label{fig:method}
\end{figure}

\section{Method}

\subsection{Problem Setup}
For a new product category, we are given $K$ normal support images $\mathcal{S}=\{\boldsymbol{x}_i^s\}_{i=1}^{K}$ without anomalous images or defect annotations, where $K$ typically set to $1$, $2$, or $4$. Given a query image $\boldsymbol{x}^q$, our goal is to predict an image-level anomaly score for detection and a pixel-level anomaly map for localization.

We first augment support images following \cite{damm2025anomalydino, zhang2023augmentation} and then employ a frozen DINOv2 \cite{oquab2023dinov2} $f_\theta(\cdot)$ to extract patch-wise features of input images, denoting the support and query features as:
\begin{equation}
\boldsymbol{f}_i^s=f_\theta(\boldsymbol{x}_i^s)\in\mathrm{R}^{N\times d},\quad
\boldsymbol{f}^q_j=f_\theta(\boldsymbol{x}^q_j)\in\mathrm{R}^{N\times d}
\label{eq:features}
\end{equation}
where $N$ is the number of patches per input image, $d$ is the dimension of each (patch-wise) feature.

\subsection{DA-SAE for Feature Reconstruction}

\paragraph{Concept Initialization.}
Concepts serve as abstract carriers that summarize the characteristics of normal samples \cite{ma2022relvit,koh2020concept}, formally denoted as $\mathcal{C}=[\boldsymbol{c}_1,\ldots,\boldsymbol{c}_M]\in \mathrm{R}^{M\times d}$,
where $M$ denotes the number of concepts. We explore two empirical strategies for concept initialization. The first leverages the available support features. In a K-shot setting with $NK$ total support features, we initialize the concepts by repeating the support features $\lfloor\frac{M}{NK}\rfloor$ times. The second alternative is random initialization.

\paragraph{Sparsemax Cross-attention.}
The core idea of our DA-SAE is to assign a sparse concepts for each input features for omitting shortcut learning, while the softmax function widely-used in Transformers usually output dense activations. Therefore, we introduce sparsemax cross-attention \cite{martins2016sparsemax} to replace the softmax function in our DA-SAE, as it is capable of producing sparse outputs. Given features extracted either from a support image or a query image, denoted by $\boldsymbol{h} \subsetneq \{\boldsymbol{f}_i^s, \boldsymbol{f}_j^q\} \in \mathbb{R}^{N \times d}$, we first feed them into a MLP-based bottleneck, and then explore a dynamic SAE with the sparsemax cross-attention, denoted by:
\begin{align}
& \boldsymbol{Q} = \boldsymbol{h} \boldsymbol{W}_Q, \ \boldsymbol{K} = \mathcal{C} \boldsymbol{W}_K, \ \boldsymbol{V} = \mathcal{C} \boldsymbol{W}_V \\
& \hat{\boldsymbol{h}} = {\rm{sparsemax}}(\boldsymbol{z}) \boldsymbol{V}, \ \boldsymbol{z} = \sigma(\frac{\boldsymbol{Q}\boldsymbol{K}^T}{\sqrt{d}}) \\
& {\rm{sparsemax}}(\boldsymbol{z}) = {\rm{argmin}}_{\boldsymbol{p} \in \Delta^{M-1}} \|\boldsymbol{p} - \boldsymbol{z}\|^2 \label{eq:sparsemax}
\end{align}
where $\sigma(\cdot)$ is softmax function, $\boldsymbol{W}_Q, \boldsymbol{W}_K, \boldsymbol{W}_V \in \mathbb{R}^{d \times d}$ denotes the query, key, and value projection matrix, respectively. ${\rm{sparsemax}}(\cdot)$ is the sparsemax function, which aims to find the similarity score inside the simplex $\Delta^{M-1}$ that is nearest to $\boldsymbol{z}$. $\Delta^{M-1}:=\{p \in \mathbb{R}^M | \boldsymbol{p}_i \ge 0, \sum_{i=1}^M \boldsymbol{p}_i = 1\}$ is the $(M-1)$-dimensional simplex. Compared with conventional softmax-activated attention, the sparsemax attention above produces exactly zero value to low-scoring concepts. Specifically, the closed-form solution of Eq. \ref{eq:sparsemax} is:
\begin{equation} \label{eq: sparemaxsolve}
{\rm{sparsemax}}(\boldsymbol{z})_m = {\rm{max}}(\boldsymbol{z}_m-\tau, 0)
\end{equation}
where $\tau$ is a threshold satisfying $\sum_{m=1}^M (\boldsymbol{z}_m - \tau) = 1$ for every $\boldsymbol{z}_m$. Hence, the indexes of selected concepts are $\mathcal{S}=\{m:\boldsymbol{z}_m > \tau\}$. Obviously, the sparsity of the selected concepts can be controlled by $\tau$ as:
\begin{align} \label{eq:taucal}
\tau = \frac{\sum_{m=1}^k \boldsymbol{z}_m - 1}{k}
\end{align}
where $k = {\rm{max}}\{n \in \{1,...,M\}|\boldsymbol{z}_n + \frac{1-\sum_{m=1}^r \boldsymbol{z}_m}{n} > 0\}$.
The threshold $\tau$ controls the sparsity of selected concepts. As shown in Fig. \ref{fig:sparsemax-tau}, rather than being a fixed hyperparameter, $\tau$ is dynamically determined by measuring the semantic complexity of the input feature $\boldsymbol{h}$. The derivations of Eq. \eqref{eq: sparemaxsolve} and \eqref{eq:taucal} can be found in the Appendix.

\paragraph{Feature Reconstruction.}
After applying sparsemax cross-attention to obtain the dynamically composed features $\hat{\boldsymbol{h}}$ from the concepts, we pass them through a norm layer and a one-layer FFN to compute the output features $\widetilde{\boldsymbol{h}}$. The concepts $\mathcal{C}$ and the DA-SAE parameters $\boldsymbol{\Phi}$ are then optimized by minimizing the feature reconstruction loss between the input and output features, formally defined as:
\begin{equation} \label{eq:fr}
\mathcal{C}^*, \boldsymbol{\Phi}^* = {\rm{argmin}}_{\mathcal{C}, \boldsymbol{\Phi}} \| \widetilde{\boldsymbol{h}} - \boldsymbol{h} \|^2
\end{equation}
where $\boldsymbol{\Phi}$ encompasses the parameters of the MLP-based bottleneck, the sparsemax attention, and the FFN.

\subsection{Adaptive Query-feature Transfer}

\paragraph{Test-time Finetuning.}
At test time, we set $\boldsymbol{h}=\boldsymbol{f}_j^q$, and employ the well-optimized DA-SAE with fixed concepts to reconstruct the input query features $\boldsymbol{h}$.
In doing so, the statistics of the query features are transferred into the trained model, mitigating the negative effects of limited representational capacity under low-data regimes. However, we intend to transfer only normal patterns while avoiding abnormal ones, yet the identity of each pattern is unknown a priori. Consequently, directly applying Eq. \eqref{eq:fr} may introduce interference from anomalies. To address this, we leverage the prior information encoded in the anomaly score. Specifically, we first compute the anomaly score map $\boldsymbol{S}=[s_1,...,s_N]$ by:
\begin{equation} \label{eq:sn}
    s_n = \|\widetilde{\boldsymbol{h}}_n - \boldsymbol{h}_n\|_2^2, \quad n=1,...,N
\end{equation}
We then sort $\boldsymbol{S}$ in descending order and select the Top-K entries with the smallest errors, assigning them a hard weight of 1 while setting the remaining entries to 0, yielding a binary weight matrix $\boldsymbol{W}=[w_1,...,w_N]$. Finally, we minimize the following weighted feature reconstruction loss:
\begin{equation} \label{eq:phiffn}
\boldsymbol{\Phi}_{FFN}^* = {\rm{argmin}}_{\boldsymbol{\Phi}_{FFN}} \sum_{n=1}^N w_n s_n, \ w_n \in [0, 1]
\end{equation}

\begin{figure}[t]
\centering
\begin{minipage}[t]{0.48\textwidth}
\vspace{0pt}
\centering
\includegraphics[width=\linewidth]{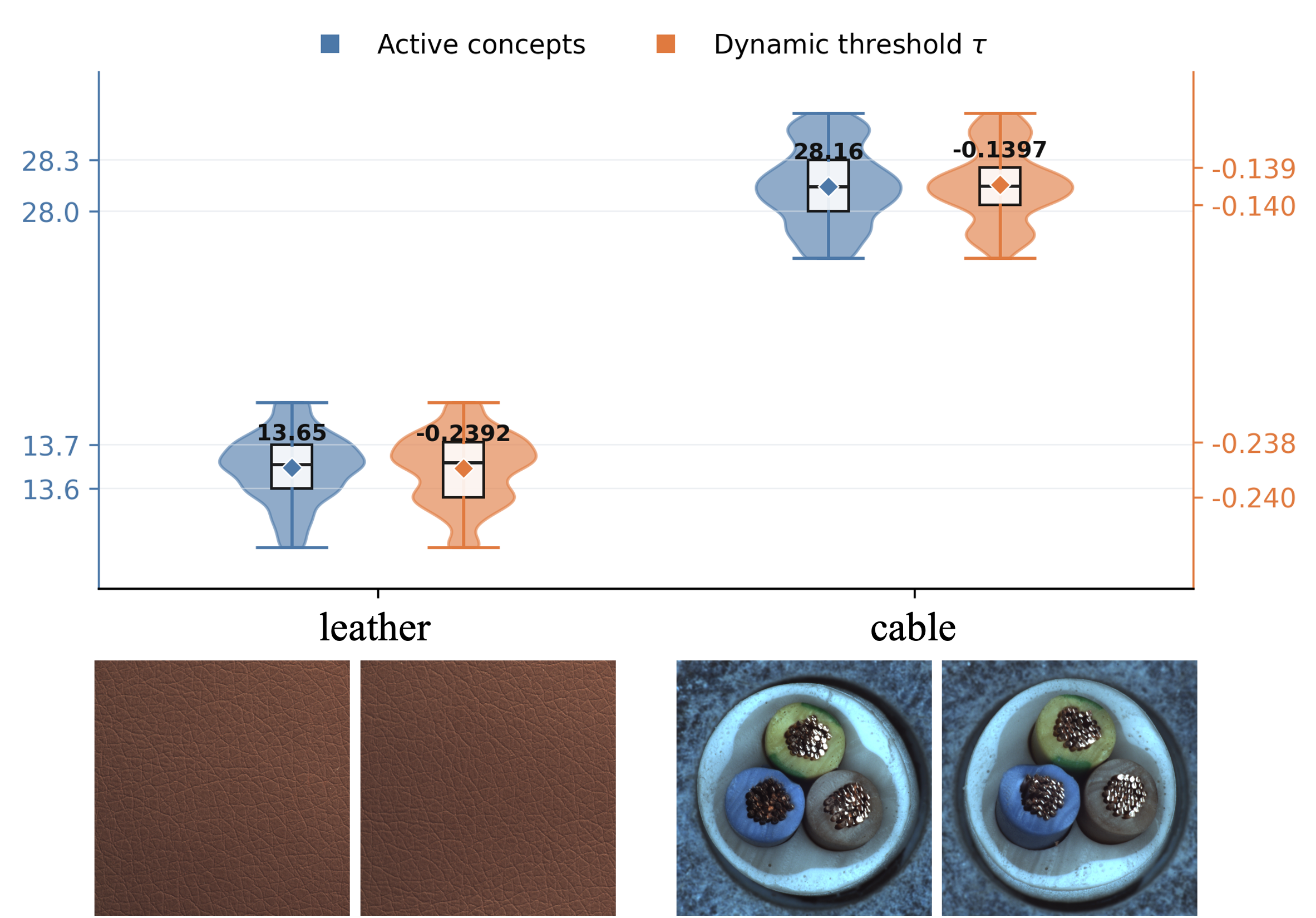}
\caption{The number of activated concepts $\mathcal{C}$ (blue) and the dynamic threshold $\tau$ (orange) on a simple category (leather) and a complex category (cable). The simple category exhibits a lower average $\tau$, which leads to a smaller number of activated $\mathcal{C}$. In contrast, the complex category has a higher average $\tau$, resulting in a larger set of activated $\mathcal{C}$.}
\label{fig:sparsemax-tau}
\end{minipage}\hfill
\begin{minipage}[t]{0.48\textwidth}
\vspace{8pt}
\refstepcounter{algorithm}
\scriptsize
\hrule
\vspace{4pt}
\textbf{Alg \thealgorithm: \method{} for FS-IAD}\\[1pt]
\raggedright
\textbf{Input:}\\[1pt]
\hspace*{1em}%
\parbox[t]{\dimexpr\linewidth-1em\relax}{%
    $K$ support images $\boldsymbol{x}^s_i$ for $i=1,\ldots,K$,\\
    the $j$-th query image $\boldsymbol{x}^q_j$,\\
    pre-trained image encoder $f_{\boldsymbol{\theta}}$,\\
    finetuning steps $T$,\\
    the number of ones $k$ in the weight matrix $\boldsymbol{W}$%
}\\[2pt]
\textbf{Parameter:}\\[1pt]
\hspace*{1em}%
\parbox[t]{\dimexpr\linewidth-1em\relax}{%
    Normal concepts $C$,\\
    DA-SAE parameters $\boldsymbol{\Phi}$%
}\\[2pt]
\textbf{Output:}\\[1pt]
\hspace*{1em}%
\parbox[t]{\dimexpr\linewidth-1em\relax}{%
    Anomaly score map $\boldsymbol{S}$%
}
\vspace{0.5pt}
\hrule
\vspace{4pt}
\begin{tabular}{@{}r p{0.82\linewidth}@{}}
1: & Extract features $\{\boldsymbol{f}_i^s\}_{i=1}^K$ and $\boldsymbol{f}_j^q$ using Eq. \eqref{eq:features}.\\
2: & Initialize normal concepts $C$ from $\{\boldsymbol{f}_i^s\}_{i=1}^K$ or randomly.\\
3: & Feed $\{\boldsymbol{f}_i^s\}_{i=1}^K$ into MLP bottleneck and DA-SAE (sparsemax cross-attention layer, norm layer, and FFN layer).\\
4: & Optimize $C$ and $\boldsymbol{\Phi}$ on $\{\boldsymbol{f}_i^s\}_{i=1}^K$ by minimizing Eq. \eqref{eq:fr}.\\
5: & \textbf{for} $t=1$ to $T$ \textbf{do}\\
6: & \quad Feed $\boldsymbol{f}_j^q$ into the MLP bottleneck and DA-SAE.\\
7: & \quad Update $\boldsymbol{\Phi}_{FFN}$ according to Eq. \eqref{eq:sn} and \eqref{eq:phiffn}.\\
8: & \textbf{end for}\\
9: & Feed $\boldsymbol{f}_j^q$ into the MLP bottleneck and DA-SAE.\\
13: & Compute anomaly score map $\boldsymbol{S}$ by Eq. \eqref{eq:sn}.\\
14: & \textbf{return} $\boldsymbol{S}$.
\end{tabular}
\vspace{4pt}
\hrule
\label{alg:pseudocode}
\end{minipage}
\end{figure}

\paragraph{Anomaly Detection.}
For image-level anomaly detection, we represent the maximum score $s^*$ among all values in anomaly score map $\boldsymbol{S}$ as $s^* = {\rm{max}}_{n \in {1,...,N}} s_n$. For pixel-level localization, we first up-scale the anomaly score map $\boldsymbol{S}$ using bi-linear interpolation to match the original input resolution. We then smooth the score map using a Gaussian kernel following \cite{roth2022patchcore}. The pseudo-code is provided in Alg. \ref{alg:pseudocode}.

\paragraph{Efficiency Analysis.}
At test time, we need to calculate the forward pass
and update FFN parameters $\boldsymbol{\Phi}_{FFN} \in \mathbb{R}^{d \times d}$, the above two steps need to iterate $T$ times. The complexity of forward pass consists of MLP bottleneck $\mathcal{O}(N d^2)$, sparsemax attention $\mathcal{O}(NMd+NM\log M)$, and FFN $\mathcal{O}(N d^2)$. The complexity of updating FFN parameters is $\mathcal{O}(N d^2)$. To sum up, the total complexity can be approximated by $\mathcal{O}(TNMd+TNM\log M+TN d^2))$.

\begin{table}[!t]
\centering
\caption{Performance of anomaly detection and localization.
Results are reported in image-level (AUROC, AP) and pixel-level (AUROC, PRO) metrics (\%).
\textbf{Bold} marks the best result within the same few-shot setting.
Missing entries are marked ``--''.}

\label{tab:fsiad_full_metrics}
\small
\setlength{\tabcolsep}{3pt}
\resizebox{0.98\textwidth}{!}{%
\begin{tabular}{cl cccc cccc cccc}
\toprule
\multirow{3}{*}{Setup} & \multirow{3}{*}{Method} & \multicolumn{4}{c}{MVTec-AD} & \multicolumn{4}{c}{MPDD} & \multicolumn{4}{c}{VisA} \\
\cmidrule(lr){3-6} \cmidrule(lr){7-10} \cmidrule(lr){11-14}
& & \multicolumn{2}{c}{Image} & \multicolumn{2}{c}{Pixel} & \multicolumn{2}{c}{Image} & \multicolumn{2}{c}{Pixel} & \multicolumn{2}{c}{Image} & \multicolumn{2}{c}{Pixel} \\
\cmidrule(lr){3-4} \cmidrule(lr){5-6} \cmidrule(lr){7-8} \cmidrule(lr){9-10} \cmidrule(lr){11-12} \cmidrule(lr){13-14}
& & AUROC & AP & AUROC & PRO & AUROC & AP & AUROC & PRO & AUROC & AP & AUROC & PRO \\
\midrule
\multirow{10}{*}{1-shot}
& RegAD~\cite{regad}              & 82.9 & --   & 92.5 & --   & 60.9 & --   & 92.6 & --   & --   & --   & --   & --   \\
& GraphCore~\cite{graphcore}      & 89.9 & --   & 95.6 & --   & \textbf{84.7} & --   & 95.2 & --   & --   & --   & --   & --   \\
& FOCT~\cite{foct}                 & 87.1 & --   & 94.4 & --   & 78.9 & --   & 96.2 & --   & 84.9   & --   & 95.5   & --   \\
& FastRecon~\cite{fastrecon}       & 85.7 & 93.1   & 93.2 & 87.0   & 74.1 & 75.3   & 96.3 & 88.1   & 76.2 & 83.4   & 96.7 & 88.5   \\
& PromptAD~\cite{promptad}         & 92.9 & 97.1 & 95.1 & 87.9 & 73.1 & --   & 95.1 & --   & 86.5 & 88.4 & 96.2 & 85.1 \\
& PatchCore~\cite{roth2022patchcore}       & 84.1 & 92.2 & 92.3 & 79.7 & 71.0 & 74.8   & 96.3 & 87.5   & 71.0 & 82.8 & 96.1 & 80.5 \\
& WinCLIP~\cite{jeong2023winclip}           & 93.5 & 96.5 & 93.6 & 87.1 & 70.5 & 72.2   & 96.3 & 88.3   & 83.4 & 85.1 & 94.7 & 85.1 \\
& AnomalyDINO~\cite{damm2025anomalydino}   & 96.4 & 98.2 & 95.8 & 92.7 & 72.0 & 74.5   & 95.9 & 90.5   & 82.2 & 89.0 & 96.3 & 92.5 \\
& FastRef~\cite{li2026fastref}              & 97.0 & 98.2   & 96.5 & 93.0   & 76.7 & 75.6     & 97.1 & 90.9     & 84.6 & 89.9   & 97.2 & 93.1   \\
& SubspaceAD~\cite{subspacead}     & \textbf{97.4} & 98.6 & \textbf{96.9} & 91.9 & 74.1 & 75.2 & \textbf{97.4} & 92.4 & 91.0 & 91.2 & 97.2 & 89.0 \\
\cmidrule{2-14}
& ${\rm{\method{}}}_{w.o.\ FT}$ (ours)                 & 97.1 & 98.5 & 96.8 & 93.2 & 78.3 & 80.7 & \textbf{97.4} & 92.7 & 92.6 & 91.0 & 97.7 & 93.4 \\
& \method{} (ours)                 & \textbf{97.4} & \textbf{98.7} & 96.8 & \textbf{93.4} & 79.9 & \textbf{81.9} & \textbf{97.4} & \textbf{92.8} & \textbf{92.8} & \textbf{91.4} & \textbf{97.8} & \textbf{93.5} \\
\midrule
\multirow{10}{*}{2-shot}
& RegAD~\cite{regad}              & 85.7 & --   & 94.6 & --   & 63.4 & --   & 93.2 & --   & --   & --   & --   & --   \\
& GraphCore~\cite{graphcore}      & 91.9 & --   & 96.9 & --   & \textbf{85.4} & --   & 95.4 & --   & --   & --   & --   & --   \\
& FOCT~\cite{foct}                 & 90.5 & --   & 94.8 & --   & 82.4 & --   & 96.5 & --   & 86.3   & --   & 95.9   & --   \\
& FastRecon~\cite{fastrecon}       & 88.3 & 94.2   & 94.5 & 87.8   & 76.4 & 76.6   & 96.7 & 89.2   & 86.1 & 87.0   & 97.6 & 89.1   \\
& PromptAD~\cite{promptad}         & 93.4 & 97.9 & 95.4 & 88.5 & 80.1 & --   & 95.8 & --   & 86.7 & 90.0 & 96.5 & 85.8 \\
& PatchCore~\cite{roth2022patchcore}       & 87.1 & 93.8 & 93.3 & 82.3 & 71.4 & 75.9   & 96.5 & 89.8   & 80.0 & 84.8 & 96.9 & 82.6 \\
& WinCLIP~\cite{jeong2023winclip}           & 93.7 & 97.0 & 93.8 & 88.4 & 72.5 & 74.3   & 96.5 & 89.5   & 83.8 & 85.8 & 95.1 & 86.2 \\
& AnomalyDINO~\cite{damm2025anomalydino}   & 96.7 & 98.2 & 96.0 & 93.1 & 75.2 & 79.1   & 96.3 & 90.7   & 82.5 & 90.7 & 96.7 & 93.4 \\
& FastRef~\cite{li2026fastref}               & 97.2 & 98.5   & 96.7 & 93.3   & 78.4 & 79.8     & 97.4 & 91.5     & 84.8 & 91.2   & 97.3 & 93.7   \\
& SubspaceAD~\cite{subspacead}     & 97.4 & 98.7 & \textbf{97.2} & 92.3 & 78.0 & 77.6 & 97.4 & 92.5 & 91.7 & 91.3 & 97.3 & 89.0 \\
\cmidrule{2-14}
& ${\rm{\method{}}}_{w.o.\ FT}$ (ours)       & 97.7 & 98.9 & 97.1 & 93.5 & 80.4 & 82.3 & 97.6 & 93.4 & 93.5 & 92.0 & 98.0 & 93.9 \\
& \method{} (ours)                 & \textbf{97.9} & \textbf{99.0} & \textbf{97.2} & \textbf{93.9} & 80.5 & \textbf{82.5} & \textbf{97.8} & \textbf{93.7} & \textbf{93.7} & \textbf{92.1} & \textbf{98.1} & \textbf{94.0} \\
\midrule
\multirow{10}{*}{4-shot}
& RegAD~\cite{regad}              & 88.2 & --   & 95.8 & --   & 68.3 & --   & 93.9 & --   & --   & --   & --   & --   \\
& GraphCore~\cite{graphcore}      & 92.9 & --   & 97.4 & --   & \textbf{85.7} & --   & 95.7 & --   & --   & --   & --   & --   \\
& FOCT~\cite{foct}                 & 93.2 & --   & 96.2 & --   & 83.2 & --   & 96.7 & --   & 90.1   & --   & 96.1   & --   \\
& FastRecon~\cite{fastrecon}       & 91.3 & 95.1   & 96.1 & 89.9   & 79.7 & 78.1   & 96.9 & 92.6   & 88.2 & 87.6   & 98.0 & 90.1   \\
& PromptAD~\cite{promptad}         & 95.5 & 98.5 & 96.3 & 90.5 & 80.4 & --   & 96.2 & --   & 88.8 & 90.8 & 96.8 & 86.2 \\
& PatchCore~\cite{roth2022patchcore}       & 90.0 & 94.5 & 95.1 & 84.3 & 76.2 & 78.0   & 97.2 & 90.3   & 84.2 & 87.5 & 97.5 & 84.9 \\
& WinCLIP~\cite{jeong2023winclip}           & 95.3 & 97.3 & 94.2 & 89.0 & 75.0 & 75.8   & 96.8 & 90.1   & 84.1 & 88.8 & 95.4 & 87.6 \\
& AnomalyDINO~\cite{damm2025anomalydino}   & 97.1 & 98.7 & 96.4 & 93.4 & 80.3 & 82.2   & 96.9 & 91.6   & 86.8 & \textbf{92.9} & 97.1 & 94.1 \\
& FastRef~\cite{li2026fastref}              & 97.4 & 98.7   & 96.8 & 93.6   & 82.1 & 82.5     & 97.6 & 92.0     & 87.1 & 92.7   & 97.9 & 93.9   \\
& SubspaceAD~\cite{subspacead}     & 97.7 & 98.9 & 97.3 & 92.6 & 78.9 & 79.4 & 97.7 & 93.1 & 92.4 & 91.8 & 97.5 & 89.0 \\
\cmidrule{2-14}
& ${\rm{\method{}}}_{w.o.\ FT}$ (ours)       & 98.0 & 98.8 & 97.2 & 93.9 & 80.6 & 82.9 & \textbf{97.9} & 93.8 & 94.0 & 92.5 & \textbf{98.2} & \textbf{94.5} \\
& \method{} (ours)                 & \textbf{98.1} & \textbf{99.1} & \textbf{97.5} & \textbf{94.2} & 80.7 & \textbf{83.6} & \textbf{97.9} & \textbf{94.3} & \textbf{94.2} & 92.8 & \textbf{98.2} & \textbf{94.5} \\
\bottomrule
\end{tabular}%
}
\end{table}

\section{Experiments}

\subsection{Datasets}
\label{sec:datasets}

We evaluate \method{} on three standard industrial anomaly detection benchmarks: MVTec-AD~\cite{mvtec}, VisA~\cite{visa}, and MPDD~\cite{mpdd}. MVTec-AD comprises 15 object and texture categories, with image resolutions ranging from 
$700 \times 700$ to $1024\times1024$. VisA contains high-resolution images of $1500 \times 1000$, featuring visually complex defects. MPDD consists of six metal-part categories with substantial structural variations and a resolution of $1024 \times 1024$.

\subsection{Evaluation Metrics and Competitive Baselines}
\label{sec:metrics}

Following standard practice~\cite{damm2025anomalydino,roth2022patchcore}, we report image-level AUROC and average precision (AP) for anomaly detection, and pixel-level AUROC and per-region overlap (PRO)~\cite{mvtec} for localization. PRO complements pixel AUROC by measuring whether each defective region is spatially covered rather than being dominated by background pixels.

We compare \method{} with recently developed FS-IAD methods including RegAD \cite{regad}, GraphCore \cite{graphcore}, FOCT \cite{foct}, PatchCore~\cite{roth2022patchcore}, AnomalyDINO~\cite{damm2025anomalydino}, SubspaceAD~\cite{subspacead}, FastRecon \cite{fastrecon}, WinCLIP~\cite{jeong2023winclip}, PromptAD~\cite{promptad}, and FastRef \cite{li2026fastref}.

\subsection{Implementation Details}
\label{sec:impl}

We adopt a frozen DINOv2-Reg ViT-B/14~\cite{oquab2023dinov2} as our backbone, extracting mean-pooled features from layers $-4$ and $-5$. All input images are center-cropped and resized to $448{\times}448$. 
For a fair comparison, we reproduce SubspaceAD using its official code, with the same backbone, input resolution, and center-crop preprocessing.
During training, for each category, we sample $k\in\{1,2,4\}$ normal support images and optimize the concepts $\mathcal{C}$ and DA-SAE parameters $\boldsymbol{\Phi}$ for up to 150 iterations, with online data augmentation using rotations at $45^{\circ}$ intervals from $0^{\circ}$ to $315^{\circ}$. 
At test time, the concepts $\mathcal{C}$ are fixed, and only the FFN parameters $\boldsymbol{\Phi}_{FFN}$ are adapted. The number of adaptation steps is set to 5, 10, and 15 for benchmarks of MVTec-AD, MPDD, and VisA, respectively.
Following prior protocols \cite{damm2025anomalydino}, we report the average results over 5 randomly sampled support images. All experiments are conducted on a single NVIDIA RTX 5880 Ada GPU.

\begin{figure}[!t]
\centering
\includegraphics[width=1.\textwidth]{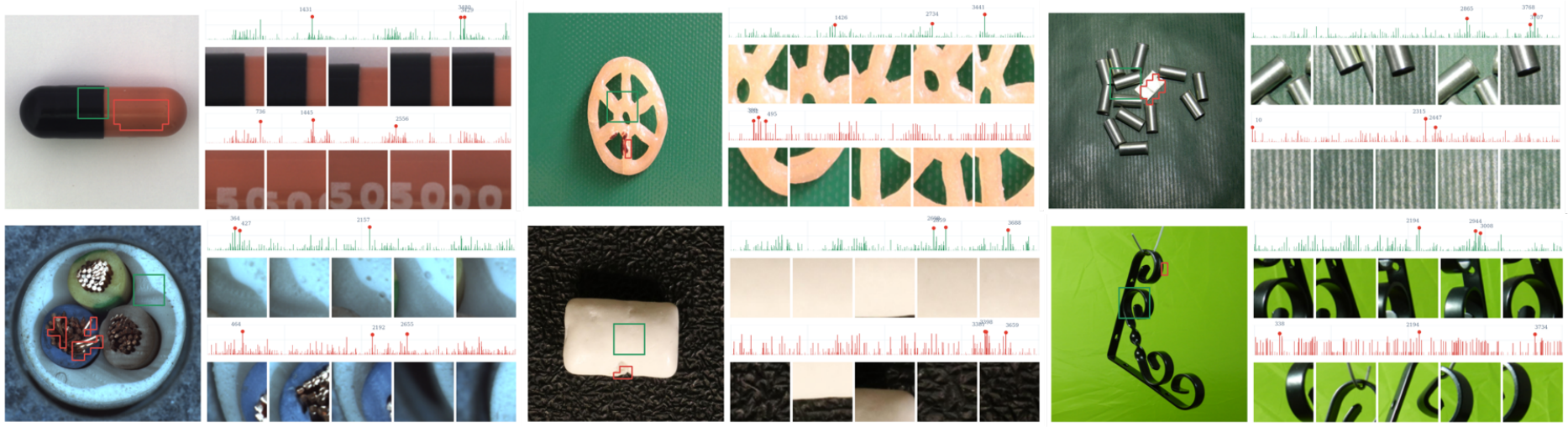}
\caption{Visualizations of the activated concepts, corresponding to normal and abnormal patches, for a given input query image. From left to right, the results correspond to the MVTec-AD, VisA, and MPDD benchmarks under 4-shot setting, respectively. For each concept, we provide its masking region with green (normal) and red (abnormal) boxes within the input query image (left), top-five most relevant patches from the support images (right), and the histogram of concept activation counts (top).}
\label{fig:concepsvis}
\end{figure}

\subsection{Comparison to SOTA Methods}
\label{sec:ourmodel_sota}
Table~\ref{tab:fsiad_full_metrics} presents a comprehensive comparison of our model against competing FS-IAD methods under various few-shot settings on standard benchmarks. To isolate the contribution of our training-stage design, we include the variant ${\rm{\method{}}}_{w.o.\ FT}$, which omits inference-time fine-tuning of the FFN parameters. Notably, this variant already outperforms non-adaptive competitors, including RegAD, GraphCore, PatchCore, and AnomalyDINO, demonstrating that our concept-guided DA-SAE architecture, trained with feature reconstruction, is inherently effective for anomaly detection and localization in low-data regimes. Even against AnomalyDINO, which shares the same foundation-model backbone, our model achieves consistent improvements across all metrics and shot settings.
Beyond foundation-model-based methods, this variant also surpasses multi-modality approaches such as WinCLIP and PromptAD, suggesting that mining domain knowledge via concepts and sparse coding offers a substantial advantage in FS-IAD. Moreover, it outperforms FOCT and FastRecon, two representative methods that enhance prototypes by transferring query feature statistics, indicating that leveraging foundation-model features for concept learning is considerably more beneficial than simply increasing the amount of training data.

When comparing our full model \method{} with FastRef, which also refines prototypes via foundation-model query feature transfer, we observe consistent improvements in most cases. We attribute this to two factors, (i) the well-optimized normal concepts and sparse reconstruction established during training provide a robust representational foundation for anomaly discrimination; and (ii) the lightweight test-time fine-tuning delivers an additional performance lift. Similarly, our method outperforms SubspaceAD, owing to the knowledge transfer enabled by our inference-time adaptation.
Notably, our model achieves substantially larger gains on more challenging benchmarks such as VisA, which features multiple foreground objects per image and subtle, hard-to-discern anomalies. For example, compared with SubspaceAD, our method boosts PRO from 89.0 \% to 94.5 \%, where a higher PRO indicates more complete defect-region coverage even when detection performance approaches saturation. 
We present qualitative visualizations in Fig.~\ref{fig:qualitative_all}. More examples are provided in the Appendix.

\subsection{Mechanistic Analysis}
To verify that our concept-guided DA-SAE effectively mitigates the shortcut learning, we present a mechanistic visualization in Fig. \ref{fig:concepsvis}. Taking the capsule category from MVTec-AD as an example, we first extract features from a normal region (marked by a green box) and retrieve the most relevant concepts from $\mathcal{C}$. We then use these selected concepts to identify the top-five most similar patches in the support images. The textures of these retrieved patches consistently reflect the morphology at the boundary of the capsule, which closely resembles the masked normal region. We then apply the same procedure to an abnormal region (marked by a red box). In this case, the retrieved patches exhibit visual appearances that are distinctly different from the abnormal region. This contrast suggests that, given a query image with anomalies, normal patches can be well reconstructed while abnormal ones cannot, thereby effectively avoiding shortcut learning. Furthermore, we observe that the activated concepts are sparse, which facilitates the identification of the most relevant concepts while avoiding unnecessary redundancy. More examples are provided in the Appendix.

\begin{figure}[t]
\centering
\begin{minipage}[t]{0.485\textwidth}
\vspace{0pt}
\centering
\includegraphics[width=\linewidth]{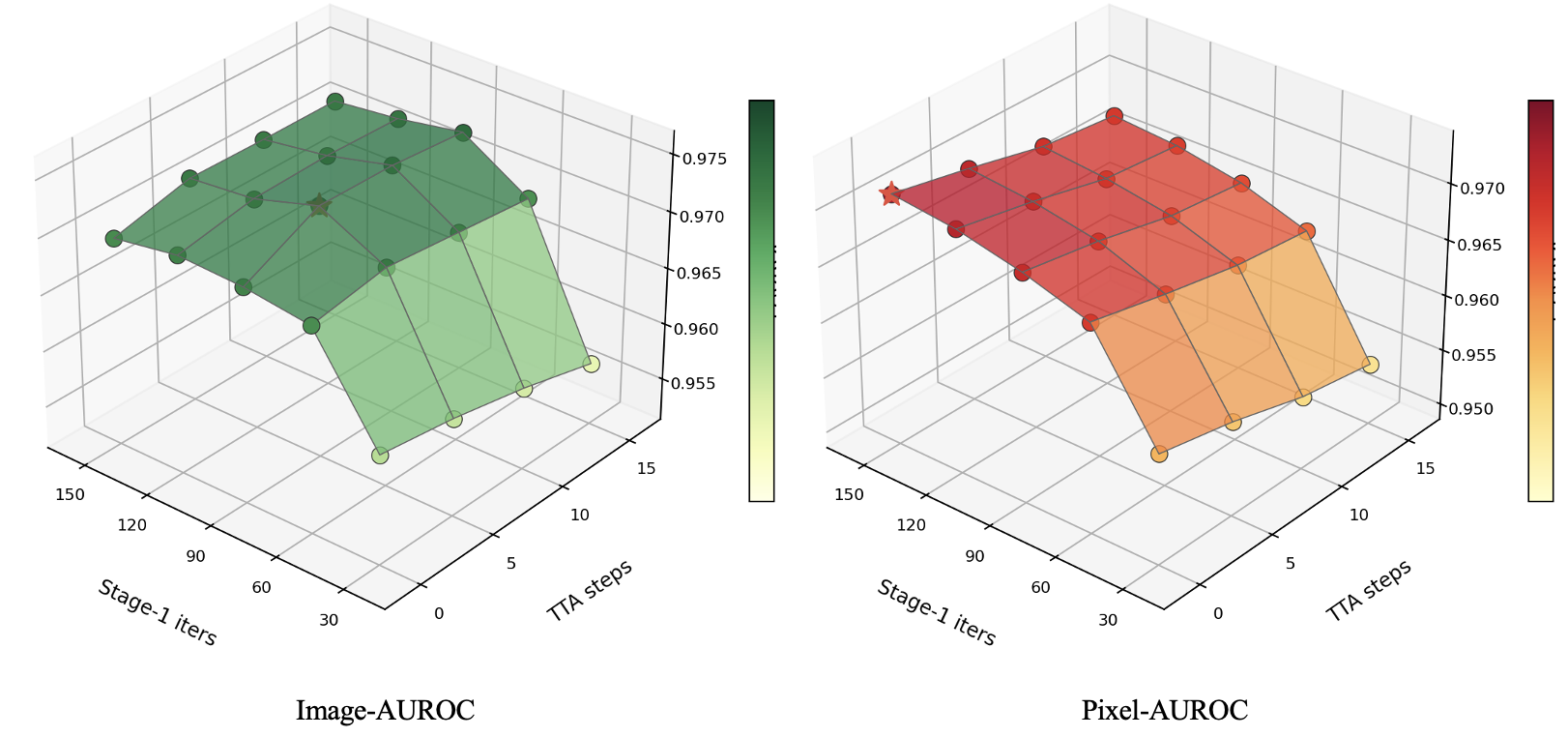}
\caption{Joint sweep of training and finetuning steps on MVTec-AD under 1-shot. The stars mark the best results.}
\label{fig:ablation_iter}

\vspace{0.5cm}

\includegraphics[width=\linewidth]{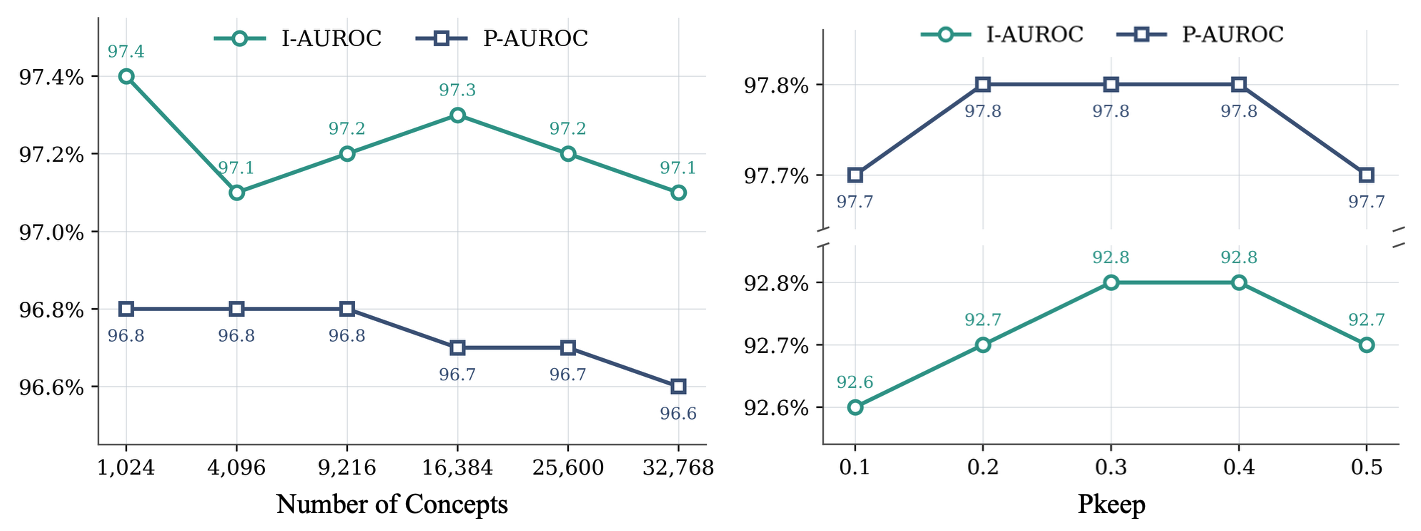}
\caption{Ablations of the number of concepts on MVTec-AD (left) and the retained proportion Pkeep of Top-K smallest-error query tokens on VisA (right) under the 1-shot setting.}
\label{fig:ablation_num_pkeep}
\end{minipage}\hfill
\begin{minipage}[t]{0.485\textwidth}
\vspace{0pt}
\centering
\tablecaption{Performance (\%) versus components on VisA.}
\vspace{5pt}
\label{tab:ablation_tta_groups}
\scriptsize
\setlength{\tabcolsep}{2.5pt}
\resizebox{\linewidth}{!}{%
\begin{tabular}{lccc}
\toprule
Method & $k{=}1$ & $k{=}2$ & $k{=}4$ \\
\midrule
\method{} & 92.6 / 97.7 & 93.5 / 98.0 & 94.0 / \textbf{98.2} \\
+ Update \texttt{001} & \textbf{92.8} / \textbf{97.8} & \textbf{93.7} / \textbf{98.1} & \textbf{94.2} / \textbf{98.2} \\
+ Update \texttt{011} & 92.7 / 97.7 & 93.6 / 97.9 & 93.9 / 98.1 \\
+ Update \texttt{111} & 92.5 / 97.5 & 93.4 / 97.8 & 93.8 / 98.0 \\
\bottomrule
\end{tabular}%
}

\tablecaption{Computational overhead with resolution $448{\times}448$.}
\vspace{5pt}
\label{tab:inference_time}
\scriptsize
\setlength{\tabcolsep}{3pt}
\resizebox{\linewidth}{!}{%
\begin{tabular}{lccc}
\toprule
Method & Backbone & Param (M) & Time (ms) \\
\midrule
WinCLIP & CLIP & 150.0 & 169 \\
AnomalyDINO & DINOv2-B & 86 & 113 \\
SubspaceAD & DINOv2-B & 86 & 56 \\
\method{} w/o FT & DINOv2-B & 86 & 118 \\
\method{} w/ FT & DINOv2-B & 86 & 145 \\
\bottomrule
\end{tabular}%
}

\tablecaption{Performance (\%) versus backbones on VisA.}
\vspace{5pt}
\label{tab:ablation_backbone}
\scriptsize
\setlength{\tabcolsep}{3pt}
\resizebox{\linewidth}{!}{%
\begin{tabular}{lccc}
\toprule
Backbone & $k{=}1$ & $k{=}2$ & $k{=}4$ \\
\midrule
EffifientNet-b4 & 66.9 / 78.5 & 67.3 / 80.1 & 67.8 / 81.1 \\
WRN50 & 76.6 / 91.4 & 80.6 / 92.4 & 81.8 / 93.8 \\
DINOv2-B & 92.8 / 97.8 & 93.7 / 98.1 & 94.2 / 98.2 \\
DINOv2-L & 94.6 / 98.1 & 95.4 / 98.3 & 95.5 / 98.5 \\
DINOv2-G & 94.7 / 98.1 & 95.6 / 98.4 & 95.7 / 98.6 \\
\bottomrule
\end{tabular}%
}
\end{minipage}
\end{figure}

\begin{figure}[t]
\centering
\includegraphics[width=0.32\textwidth]{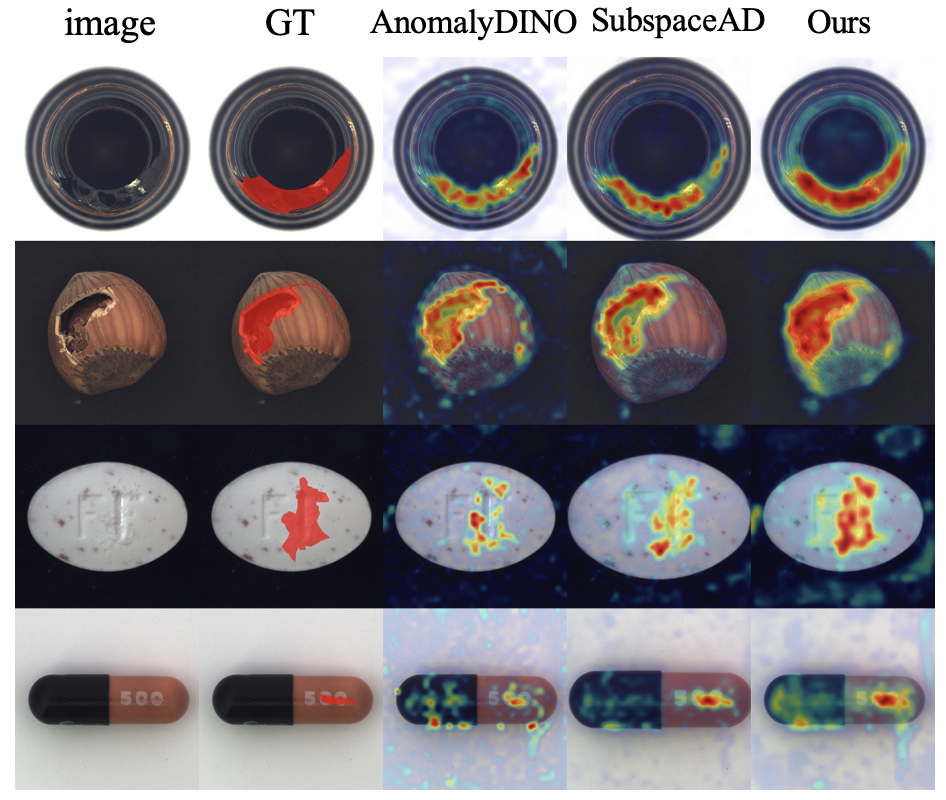}\hfill
\includegraphics[width=0.32\textwidth]{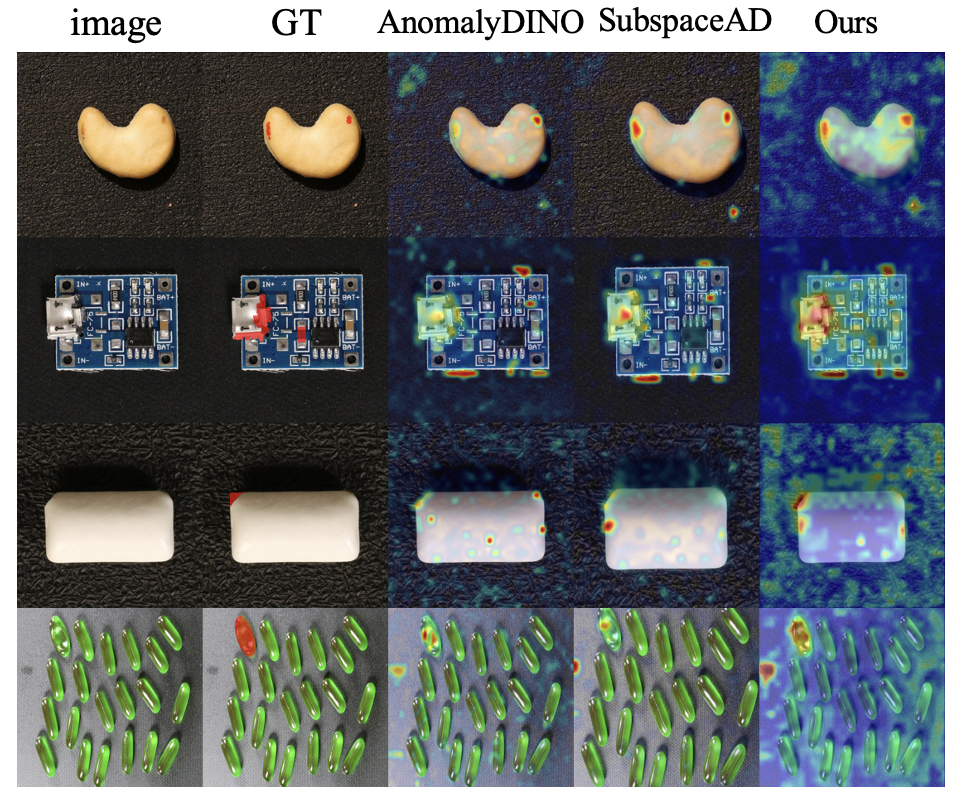}\hfill
\includegraphics[width=0.32\textwidth]{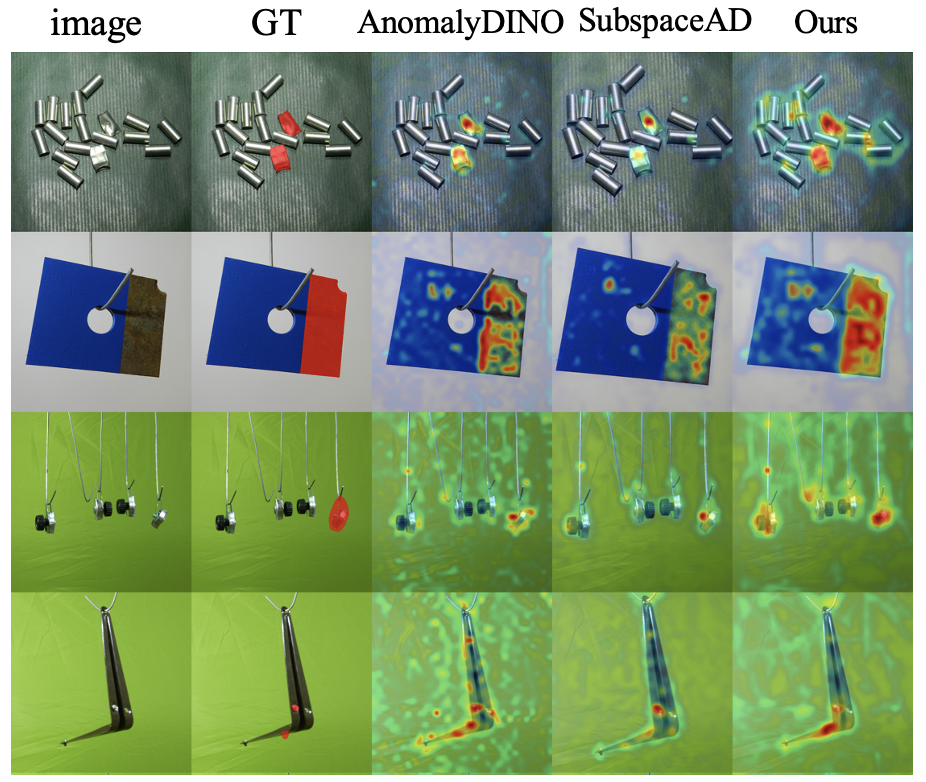}
\caption{Qualitative comparison on benchmarks of MVTec-AD (left), VisA (center), and MPDD (right) under 1-shot setting.
}
\label{fig:qualitative_all}

\end{figure}

\subsection{Ablation Study}
\label{sec:ourmodel_ablation}

\paragraph{Training and Finetuning Steps.}
In Table~\ref{tab:ablation_tta_groups}, the three binary bits indicate whether the concepts, sparsemax attention, and FFN components are updated during inference. Updating only the FFN (\texttt{001}) achieves the best performance while maintaining a lightweight design, whereas adapting the other two components leads to a slight performance drop. This confirms that updating the FFN alone provides sufficient flexibility for query-feature transfer, making it advantageous for fast adaptation in terms of parameter efficiency.
Figure~\ref{fig:ablation_iter} shows that performance first improves and then declines as the number of training steps increases, with a rapid drop occurring within the first few fine-tuning steps. Excessive adaptation causes a performance degradation, suggesting that conservative updates are more effective at avoiding the transfer of anomalous patterns. This observation further supports the benefit of fast adaptation from a time-efficiency perspective.

\paragraph{Concept Size and Top-K Entries.}
Figure~\ref{fig:ablation_num_pkeep} (left) shows that increasing the number of concepts yields only marginal gains in both detection and localization metrics. This suggests that an excessive number of concepts may increase learning difficulty given limited support images, and thus a lightweight concept set is preferable. Meanwhile, the right panel indicates that retaining 30 $\sim$ 40 \% of the query features with the smallest reconstruction errors as Top-K entries achieves the best performance. This is because too few features fail to provide sufficient normal statistics, while a larger proportion may introduce anomalous patterns.

\paragraph{Backbone Generalization.}
We report performance versus backbones from CNN-based  models to Transformer-based foundation models in table~\ref{tab:ablation_backbone}. It confirms that foundation-model features facilitate the enhanced anomaly detection.

\paragraph{Computational Overhead.}
Table~\ref{tab:inference_time} reports the computational overhead. Our model incurs slightly higher time costs than SubspaceAD, primarily due to the sparsemax attention computation and the additional FFN updates during inference. Overall, however, our model maintains competitive computational overhead compared with the baselines.

\section{Conclusion}

We present a concept-guided adaptive feature reconstruction model for FS-IAD. Our model learns normal concepts via sparse reconstruction, mitigating shortcut learning with limited data. At inference, we fix the concepts and update the FFN layer using reliable query features, enabling query adaptation without prototype shift. This design achieves both stability and flexibility from few support images. Experiments on three benchmarks show consistent gains in detection and localization, and mechanistic analyses validate our design. Overall, decoupling concept learning from query adaptation offers an effective solution for low-data industrial inspection.

\bibliographystyle{plainnat}
\bibliography{main}

@inproceedings{roth2022patchcore,
  author    = {Roth, Karsten and Pemula, Latha and Zepeda, Joaquin and Sch{\"o}lkopf, Bernhard and Brox, Thomas and Gehler, Peter},
  title     = {Towards Total Recall in Industrial Anomaly Detection},
  booktitle = {Proceedings of the IEEE/CVF Conference on Computer Vision and Pattern Recognition},
  pages     = {14318--14328},
  year      = {2022}
}

@inproceedings{jeong2023winclip,
  author    = {Jeong, Jongheon and Zou, Yang and Kim, Taewan and Zhang, Dongqing and Ravichandran, Avinash and Dabeer, Onkar},
  title     = {WinCLIP: Zero-/Few-Shot Anomaly Classification and Segmentation},
  booktitle = {Proceedings of the IEEE/CVF Conference on Computer Vision and Pattern Recognition},
  pages     = {19606--19616},
  year      = {2023}
}

@article{santos2023fspatchcore,
  author  = {Santos, Jo{\~a}o and Tran, Triet and Rippel, Oliver},
  title   = {Optimizing PatchCore for Few/many-shot Anomaly Detection},
  journal = {arXiv preprint arXiv:2307.10792},
  year    = {2023}
}

@inproceedings{damm2025anomalydino,
  author    = {Damm, Simon and Laszkiewicz, Mike and Lederer, Johannes and Fischer, Asja},
  title     = {AnomalyDINO: Boosting Patch-Based Few-Shot Anomaly Detection with DINOv2},
  booktitle = {Proceedings of the IEEE/CVF Winter Conference on Applications of Computer Vision},
  pages     = {1319--1329},
  year      = {2025}
}

@inproceedings{deng2024dinomaly,
  author    = {Guo, Jia and Lu, Shuai and Zhang, Weihang and Chen, Fang and Li, Huiqi and Liao, Hongen},
  title     = {Dinomaly: The Less Is More Philosophy in Multi-Class Unsupervised Anomaly Detection},
  booktitle = {Proceedings of the IEEE/CVF Conference on Computer Vision and Pattern Recognition},
  year      = {2025}
}

@inproceedings{li2026fastref,
  author    = {Li, Yufei and Tian, Long and Dai, Yuyang and Chen, Wenchao and Bao, Liang and Liu, Xiyang},
  title     = {FastRef: Fast Prototype Refinement for Few-Shot Industrial Anomaly Detection},
  booktitle = {Proceedings of the IEEE/CVF Conference on Computer Vision and Pattern Recognition},
  year      = {2026}
}

@inproceedings{ma2022relvit,
  author    = {Ma, Xiaojian and Nie, Weili and Yu, Zhiding and Jiang, Huaizu and Xiao, Chaowei and Zhu, Yuke and Zhu, Song-Chun and Anandkumar, Anima},
  title     = {RelViT: Concept-Guided Vision Transformer for Visual Relational Reasoning},
  booktitle = {International Conference on Learning Representations},
  year      = {2022}
}

@inproceedings{martins2016sparsemax,
  author    = {Martins, Andr{\'e} F. T.},
  title     = {From Softmax to Sparsemax: A Sparse Model of Attention and Multi-Label Classification},
  booktitle = {Proceedings of the 33rd International Conference on Machine Learning},
  pages     = {1614--1623},
  year      = {2016}
}

@inproceedings{he2016deep,
  title={Deep residual learning for image recognition},
  author={He, Kaiming and Zhang, Xiangyu and Ren, Shaoqing and Sun, Jian},
  booktitle={Proceedings of the IEEE conference on computer vision and pattern recognition},
  pages={770--778},
  year={2016}
}

@inproceedings{tan2019efficientnet,
  title={Efficientnet: Rethinking model scaling for convolutional neural networks},
  author={Tan, Mingxing and Le, Quoc},
  booktitle={International conference on machine learning},
  pages={6105--6114},
  year={2019},
  organization={PMLR}
}

@inproceedings{radford2021learning,
  title={Learning transferable visual models from natural language supervision},
  author={Radford, Alec and Kim, Jong Wook and Hallacy, Chris and Ramesh, Aditya and Goh, Gabriel and Agarwal, Sandhini and Sastry, Girish and Askell, Amanda and Mishkin, Pamela and Clark, Jack and others},
  booktitle={International conference on machine learning},
  pages={8748--8763},
  year={2021},
  organization={PmLR}
}

@inproceedings{hvqtrans,
author = {Lu, Ruiying and Wu, YuJie and Tian, Long and Wang, Dongsheng and Chen, Bo and Liu, Xiyang and Hu, Ruimin},
title = {Hierarchical vector quantized transformer for multi-class unsupervised anomaly detection},
year = {2023},
booktitle = {Proceedings of the 37th International Conference on Neural Information Processing Systems},
articleno = {370},
numpages = {14},
location = {New Orleans, LA, USA},
series = {NIPS '23}
}

@inproceedings{oquab2023dinov2,
  author    = {Oquab, Maxime and Darcet, Timoth{\'e}e and Moutakanni, Th{\'e}o and Vo, Huy V. and Szafraniec, Marc and Khalidov, Vasil and Fernandez, Pierre and Haziza, Daniel and Massa, Francisco and El-Nouby, Alaaeldin and Assran, Mahmoud and Ballas, Nicolas and Galuba, Wojciech and Howes, Russell and Huang, Po-Yao and Li, Shang-Wen and Misra, Ishan and Rabbat, Michael and Sharma, Vasu and Synnaeve, Gabriel and Xu, Hu and J{\'e}gou, Herv{\'e} and Mairal, Julien and Labatut, Patrick and Joulin, Armand and Bojanowski, Piotr},
  title     = {DINOv2: Learning Robust Visual Features without Supervision},
  booktitle = {Transactions on Machine Learning Research},
  year      = {2024}
}

@inproceedings{mvtec,
  title={{MVTec AD}--A Comprehensive Real-World Dataset for Unsupervised Anomaly Detection},
  author={Bergmann, Paul and Fauser, Michael and Sattlegger, David and Steger, Carsten},
  booktitle={CVPR},
  year={2019}
}

@inproceedings{visa,
  title={Spot-the-Difference Self-Supervised Pre-training for Anomaly Detection and Segmentation},
  author={Zou, Yang and others},
  booktitle={ECCV},
  year={2022}
}

@inproceedings{mpdd,
  title={{MPDD}: A Multi-Part Industrial Image Dataset for Anomaly Detection},
  author={Jiang, Yuxin and others},
  booktitle={arXiv},
  year={2023}
}

@inproceedings{subspacead,
  title={{SubspaceAD}: Training-Free Few-Shot Anomaly Detection via Subspace Modeling},
  author={Lendering, Camile and others},
  booktitle={CVPR},
  year={2026}
}

@inproceedings{promptad,
  title={{PromptAD}: Learning Prompts with only Normal Samples for Few-Shot Anomaly Detection and Segmentation},
  author={Li, Guang and others},
  booktitle={CVPR},
  year={2024}
}

@article{hu2022lora,
  title={Lora: Low-rank adaptation of large language models.},
  author={Hu, Edward J and Shen, Yelong and Wallis, Phillip and Allen-Zhu, Zeyuan and Li, Yuanzhi and Wang, Shean and Wang, Liang and Chen, Weizhu and others},
  journal={Iclr},
  volume={1},
  number={2},
  pages={3},
  year={2022}
}

@article{you2022unified,
  title={A unified model for multi-class anomaly detection},
  author={You, Zhiyuan and Cui, Lei and Shen, Yujun and Yang, Kai and Lu, Xin and Zheng, Yu and Le, Xinyi},
  journal={Advances in Neural Information Processing Systems},
  volume={35},
  pages={4571--4584},
  year={2022}
}

@article{cuturi2013sinkhorn,
  title={Sinkhorn distances: Lightspeed computation of optimal transport},
  author={Cuturi, Marco},
  journal={Advances in neural information processing systems},
  volume={26},
  year={2013}
}

@inproceedings{wu2021learning,
  title={Learning unsupervised metaformer for anomaly detection},
  author={Wu, Jhih-Ciang and Chen, Ding-Jie and Fuh, Chiou-Shann and Liu, Tyng-Luh},
  booktitle={Proceedings of the IEEE/CVF international conference on computer vision},
  pages={4369--4378},
  year={2021}
}

@article{agarwal2005geometric,
  title={Geometric approximation via coresets},
  author={Agarwal, Pankaj K and Har-Peled, Sariel and Varadarajan, Kasturi R and others},
  journal={Combinatorial and computational geometry},
  volume={52},
  number={1},
  pages={1--30},
  year={2005}
}

@inproceedings{regad,
  author    = {Huang, Chaoqin and Guan, Haoyan and Jiang, Aofan and Zhang, Ya and Spratling, Michael and Wang, Yan-Feng},
  title     = {Registration Based Few-Shot Anomaly Detection},
  booktitle = {European Conference on Computer Vision},
  pages     = {303--319},
  year      = {2022}
}

@inproceedings{graphcore,
  author    = {Xie, Guoyang and Wang, Jinbao and Liu, Jiaqi and Zheng, Feng and Jin, Yaochu},
  title     = {Pushing the Limits of Few-Shot Anomaly Detection in Industry Vision: {GraphCore}},
  booktitle = {International Conference on Learning Representations},
  year      = {2023}
}

@inproceedings{fastrecon,
  author    = {Fang, Zheng and Wang, Xiaoyang and Li, Haocheng and Liu, Jiejie and Hu, Qiugui and Xiao, Jimin},
  title     = {{FastRecon}: Few-Shot Industrial Anomaly Detection via Fast Feature Reconstruction},
  booktitle = {Proceedings of the IEEE/CVF International Conference on Computer Vision},
  pages     = {17481--17490},
  year      = {2023}
}

@inproceedings{foct,
  author    = {Tian, Long and Zhao, Hongyi and Lu, Ruiying and Wang, Rongrong and Wu, Yujie and Wang, Liming and He, Xiongpeng and Liu, Xiyang},
  title     = {{FOCT}: Few-Shot Industrial Anomaly Detection with Foreground-Aware Online Conditional Transport},
  booktitle = {Proceedings of the 32nd ACM International Conference on Multimedia},
  pages     = {6241--6249},
  year      = {2024},
  doi       = {10.1145/3664647.3680771}
}

@inproceedings{defard2021padim,
  author    = {Defard, Thomas and Setkov, Aleksandr and Loesch, Angelique and Audigier, Romaric},
  title     = {{PaDiM}: A Patch Distribution Modeling Framework for Anomaly Detection and Localization},
  booktitle = {Pattern Recognition. ICPR International Workshops and Challenges},
  series    = {Lecture Notes in Computer Science},
  volume    = {12664},
  pages     = {475--489},
  publisher = {Springer},
  year      = {2021},
  doi       = {10.1007/978-3-030-68799-1_35}
}

@inproceedings{zavrtanik2021draem,
  author    = {Zavrtanik, Vitjan and Kristan, Matej},
  title     = {{DRAEM}: A Discriminatively Trained Reconstruction Embedding for Surface Anomaly Detection},
  booktitle = {Proceedings of the IEEE/CVF International Conference on Computer Vision},
  pages     = {8330--8339},
  year      = {2021}
}

@inproceedings{deng2022reverse,
  author    = {Deng, Hanqiu and Li, Xingyu},
  title     = {Anomaly Detection via Reverse Distillation from One-Class Embedding},
  booktitle = {Proceedings of the IEEE/CVF Conference on Computer Vision and Pattern Recognition},
  pages     = {9737--9746},
  year      = {2022}
}

@inproceedings{liu2023simplenet,
  author    = {Liu, Zhikang and Zhou, Yiming and Xu, Yuansheng and Wang, Zilei},
  title     = {{SimpleNet}: A Simple Network for Image Anomaly Detection and Localization},
  booktitle = {Proceedings of the IEEE/CVF Conference on Computer Vision and Pattern Recognition},
  pages     = {20402--20411},
  year      = {2023}
}

@inproceedings{batzner2024efficientad,
  author    = {Batzner, Kilian and Heckler, Lars and K{\"o}nig, Rebecca},
  title     = {{EfficientAD}: Accurate Visual Anomaly Detection at Millisecond-Level Latencies},
  booktitle = {Proceedings of the IEEE/CVF Winter Conference on Applications of Computer Vision},
  pages     = {128--138},
  year      = {2024}
}

@inproceedings{zhang2023augmentation,
  author    = {Zhang, Lingrui and Zhang, Shuheng and Xie, Guoyang and Liu, Jiaqi and Yan, Hua and Wang, Jinbao and Zheng, Feng and Jin, Yaochu},
  title     = {What Makes a Good Data Augmentation for Few-Shot Unsupervised Image Anomaly Detection?},
  booktitle = {Proceedings of the IEEE/CVF Conference on Computer Vision and Pattern Recognition Workshops},
  pages     = {4345--4354},
  year      = {2023}
}

@inproceedings{koh2020concept,
  author    = {Koh, Pang Wei and Nguyen, Thao and Tang, Yew Siang and Mussmann, Stephen and Pierson, Emma and Kim, Been and Liang, Percy},
  title     = {Concept Bottleneck Models},
  booktitle = {Proceedings of the 37th International Conference on Machine Learning},
  series    = {Proceedings of Machine Learning Research},
  volume    = {119},
  pages     = {5338--5348},
  publisher = {PMLR},
  year      = {2020}
}

\end{document}